\pdfoutput=1

\documentclass[11pt]{article}

\usepackage[final]{acl}

\usepackage{times}
\usepackage{latexsym}
\usepackage{adjustbox}
\usepackage{makecell}
\usepackage{multirow}
\usepackage{ulem}
\usepackage[T1]{fontenc}

\usepackage[utf8]{inputenc}

\usepackage{microtype}

\usepackage{inconsolata}

\usepackage{graphicx}

\title{Agent-UniRAG: A Trainable Open-Source LLM Agent Framework for Unified Retrieval-Augmented Generation Systems}

\author{
  Hoang Pham\textsuperscript{1}, 
  Thuy-Duong Nguyen\textsuperscript{2}, 
  Khac-Hoai Nam Bui\textsuperscript{1}\thanks{Corresponding author} \\
  \textsuperscript{1}Viettel Artificial Intelligence and Data Services Center, Viettel Group \\
  \textsuperscript{2}Hanoi University of Science and Technology \\
  \{hoangpv4, nambkh\}@viettel.com.vn, duong.nt204536@sis.hust.edu.vn
}

\begin{document}
\maketitle

\begin{abstract}
This paper presents a novel approach for unified retrieval-augmented generation (RAG) systems using the recent emerging large language model (LLM) agent concept. Specifically, Agent LLM, which utilizes LLM as fundamental controllers, has become a promising approach to enable the interpretability of RAG tasks, especially for complex reasoning question-answering systems (e.g., multi-hop queries). Nonetheless, previous works mainly focus on solving RAG systems with either single-hop or multi-hop approaches separately, which limits the application of those approaches to real-world applications. In this study, we propose a trainable agent framework called Agent-UniRAG for unified retrieval-augmented LLM systems, which enhances the effectiveness and interpretability of RAG systems. The main idea is to design an LLM agent framework to solve RAG tasks step-by-step based on the complexity of the inputs, simultaneously including single-hop and multi-hop queries in an end-to-end manner. Furthermore, we introduce SynAgent-RAG, a synthetic dataset to enable the proposed agent framework for small open-source LLMs (e.g., Llama-3-8B). The results show comparable performances with closed-source and larger open-source LLMs across various RAG benchmarks. Our source code and dataset are publicly available for further exploitation.
\end{abstract}

\section{Introduction}
Incorporating non-parametric knowledge into large language models (LLMs) through additional retrieval modules has emerged as a promising approach to enhance both accuracy and the timeliness of information \cite{BorgeaudMHCRM0L22, IzacardLLHPSDJRG23}. This issue has led to the rapid development of various retrieval-augmented LLM paradigms designed to provide correct answers to user queries. Accordingly, these modern paradigms address either single-hop which can respond within a single document (i.e., Naive RAG), or complex multi-hop queries, which require the integration and synthesis of information from multiple documents (i.e., Advanced RAG)\cite{FanDNWLYCL24}.

\begin{figure}[t]
  \includegraphics[width=\columnwidth]{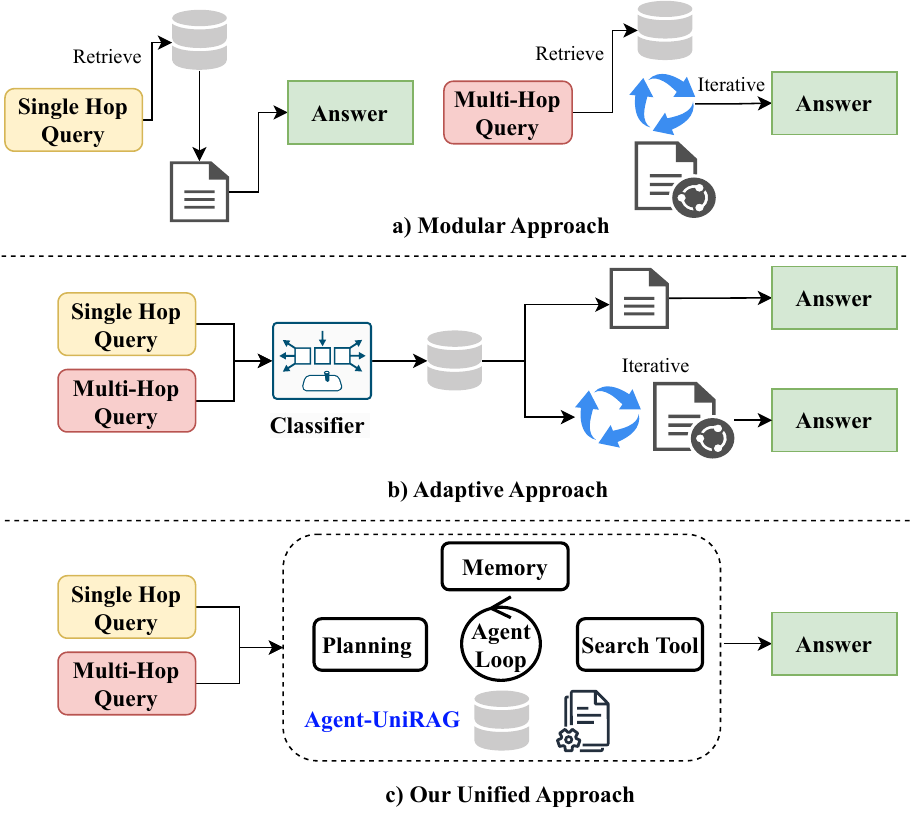}
  \caption{Conceptual analysis of previous works and Agent-UniRAG: (a) Modular approach handles query types separately; (b) Adaptive approach uses a classifier to determine query types before executing them separately; (c) Agent-UniRAG processes all query types within a unified system using the Agent LLM concept.}
  \label{fig:comparision}
\end{figure}

Nonetheless, existing modern approaches suffer from several significant limitations, including a lack of explainability and traceability. Accordingly, an emerging research issue in this regard is that current methods either inefficiently handle simple queries with unnecessary computational overhead or fail to address complex multi-step queries \cite{abs-2401-15391} (Figure \ref{fig:comparision} (a)). To address this research issue, a potential method is to add a classification module to classify the complexity of input queries for selecting the appropriate RAG model to respond \cite{JeongBCHP24} (Figure \ref{fig:comparision} (b)). However, this approach is only suitable when the types of queries are predefined (in specific domains or custom benchmark datasets), which might lack flexibility and scalability in terms of various real-world applications. Recently, LLM Agent, by leveraging LLMs to execute complex tasks, emerged as a promising approach to enable the interpretability and reasoning capability for LLM \cite{Zhao0XLLH24}.  Specifically, LLM is regarded as the primary controller, integrating with essential components such as planning, memory, and action execute operations necessary to complex tasks\cite{Wang_2024}. Based on this emerging conceptual technology, this study raises a research question: \textit{Can the LLM agent enable the interpretability and reasoning capability of RAG systems in a unified manner?}

Figure \ref{fig:comparision} (c) illustrates our proposed approach, which is designed to enhance the interpretability and effectiveness of LLMs in RAG tasks, compared with previous approaches in this research field. Specifically, we leverage the emerging concept of LLM agents, employing LLMs as central controllers to unify RAG tasks. Our unified agent is capable of handling queries that require reasoning processes (including both single-hop and multi-hop queries simultaneously) through self-guided instructions and interaction with the external knowledge base. Furthermore, most current LLM agent frameworks rely on closed-source models with very large weight sizes (e.g., GPT-4 \cite{openai2024gpt4}), which limits their reproducibility and controllability. Our primary focus, therefore, is on enabling trainable open-source LLM agents. In this regard, we also introduce a synthetic dataset named \textit{SynAgent-RAG} to train these open-source LLM-based agents for the unified RAG system. In summary, the main contributions of this study are three-fold as follows: 

\textbf{(i)} We propose a unified RAG system using the concept of the LLM agent, which can handle queries that require reasoning processes (e.g. single-hop and multi-hop queries) by self-guided instructions and interaction with the external knowledge base to derive a response to the input queries. To the best of our knowledge, this paper is the first study to execute the unified RAG system in an end-to-end manner.

\textbf{(ii)} We process and introduce the \textit{SynAgent-RAG} dataset, which obtains 16,987 synthetic samples to enable small open-source modern LLMs (e.g., Llama-3-8B) to adapt the proposed Agent-UniRAG approach via instruction finetuning. Accordingly, this contribution is important to achieve the desired flexibility and scalability since most emerging LLM Agent technologies only work well with very large LLMs as the backbone.

\textbf{(iii)} We evaluate the proposed approach on various RAG benchmarks, including the test set of our proposed SynAgent-RAG dataset. The experimental results show that our approach outperforms previous approaches. Furthermore, with small LLMs (e.g., Llama-3-8B) instruction-finetuned on the proposed dataset, we can achieve competitive performances compared to closed-source (e.g., GPT-4) and larger open-source agent LLMs (e.g., Llama-3-70B).

\section{Literature Reviews}
\subsection{Retrieval-Augmented LLM}
The evolution of RAG in the era of LLMs can be divided into three categories, including Naive RAG, Advanced RAG, and Modular RAG \cite{abs-2312-10997}. Naive RAG and Advanced RAG are typical \textit{Retrieve-Read} paradigms \cite{ma-etal-2023-query}, which focus on finding the answers in a single document (i.e., single-hop queries \cite{ram-etal-2023-context}). Meanwhile, the recent emerging Modular RAG has been introduced to go beyond the two aforementioned RAG paradigms, which requires iterative accesses to both LLMs and retrievers multiple times (i.e., multi-hop queries \cite{trivedi-etal-2023-interleaving}). Specifically, dynamically selecting the suitable strategy (i.e., single-hop or multi-hop) for unified RAG tasks become an emerging research issue in this research field \cite{JeongBCHP24}.

\subsection{LLM Agent Framework}
The concept of LLM agents involves LLM applications that can execute complex tasks, in which LLMs serve as controllers to control the flow of operations needed to complete a task or user request \cite{Wang_2024}. Accordingly, an LLM agent framework consists of the four core components such as \textit{User Request}, \textit{Agent}, \textit{Planning}, and \textit{Memory}. HuggingGPT \cite{0001ST00Z23} was introduced as one of the first comprehensive LLM-powered agent frameworks, which use LLMs (i.e., ChatGPT) and the ML community (i.e., Hugging Face) to process inputs from different modalities. Sequentially, \citet{yin2023lumos} introduces LUMOS, an agent framework for trainable open-source LLM. Specifically, the framework designs a modular architecture with a planning module to learn subgoals and a grounding module trained to translate subgoals into actions, using tools in the execution module. Inspired by previous works, in this study, we present a trainable open-source LLM-based agent framework for unified RAG tasks, which focuses on integrating the interpretable ability of LLM to determine the next action for solving RAG tasks.

\section{Methodology}
\begin{figure}[!h]
  \centering
  \includegraphics[width=\columnwidth]{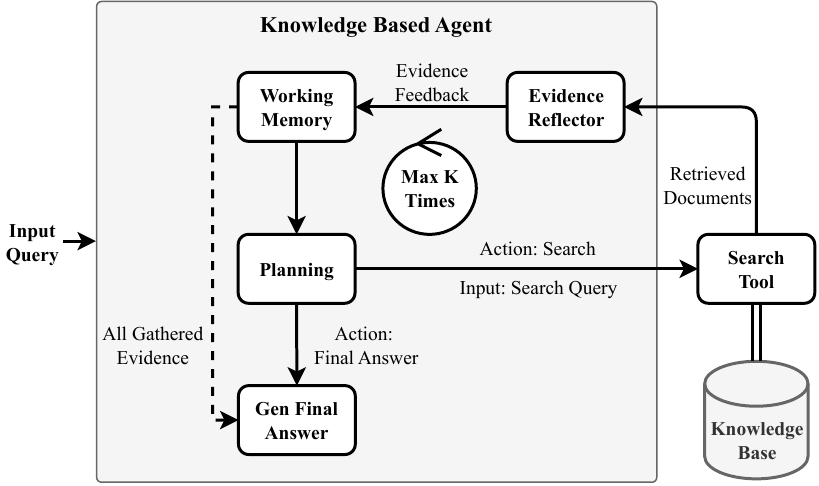}
  \caption{Overall design of Agent-UniRAG.}
  \label{fig:Agent-UniRAG}
\end{figure}

This section introduces the design of Agent-UniRAG. Following the typical pipeline of the LLM agent framework, our framework - Agent-UniRAG is put into a loop and includes four main components including Planning Module, Tool Using Module, Working Memory Module, and Reflector Module as shown in Figure \ref{fig:Agent-UniRAG}.

\subsection{Planning Module}
Leveraging the reasoning capabilities of modern LLMs, this module is designed to systematically determine the necessary actions required to address a user's request (Input Query) at each step of the process. Specifically, the agent decides between two primary actions at each decision point: 
\begin{itemize}
    \item \textbf{Action: Search} – This action is triggered when the agent needs to acquire additional external knowledge to progress toward solving the problem.
    \item \textbf{Action: Final Answer} – This action is taken when the agent has accumulated sufficient information to confidently provide a response to the query.
\end{itemize}
To implement this decision-making process, Agent-UniRAG utilizes the ReAct mechanism \cite{YaoZYDSN023}, which allows the agent to iteratively reflect on and refine its execution plan. The mechanism guides the agent through a structured sequence of steps: \textit{Thought}, \textit{Action}, and \textit{Evidence Feedback}. Continuously evaluating and integrating those steps, the agent is capable of addressing complex tasks with great precision.

\subsection{Search Tool}
At each stage where external knowledge is required (\textit{Action: Search}), the agent interacts with the Knowledge Base through the Search Tool by formulating a search query generated by the Planning Module. The purpose of querying external knowledge is to ground the reasoning process in reliable and up-to-date information beyond the agent's internal knowledge. This ensures that the agent's responses are accurate and contextually relevant, especially for tasks requiring current or specialized domain knowledge. The retrieved external evidence supports the resolution of the input query, functioning as a document retrieval task.

\subsection{Reflector Module}
Documents retrieved from external knowledge bases often include irrelevant or extraneous information, especially when the knowledge base cannot adequately satisfy the query. Incorporating such unfiltered data into LLMs can introduce noise, degrade performance, or even mislead the model. Inspired by \cite{shinn2024reflexion}, to mitigate this issue, we designed a module called Evidence Reflector to provide evidence feedback to LLM, which operates after the Search Tool. The Evidence Reflector filters out irrelevant content and refines the retrieved information, delivering back more focused and relevant insights to the agent. If no suitable evidence is found, it feedbacks with "No information found." This feedback is critical in guiding the model’s subsequent actions, ensuring the decision-making process remains both accurate and efficient. The agent can then better locate and leverage relevant information, thereby improving both the quality and precision of its responses.

\begin{figure*}[!h]
    \centering
    \includegraphics[width=0.9\textwidth]{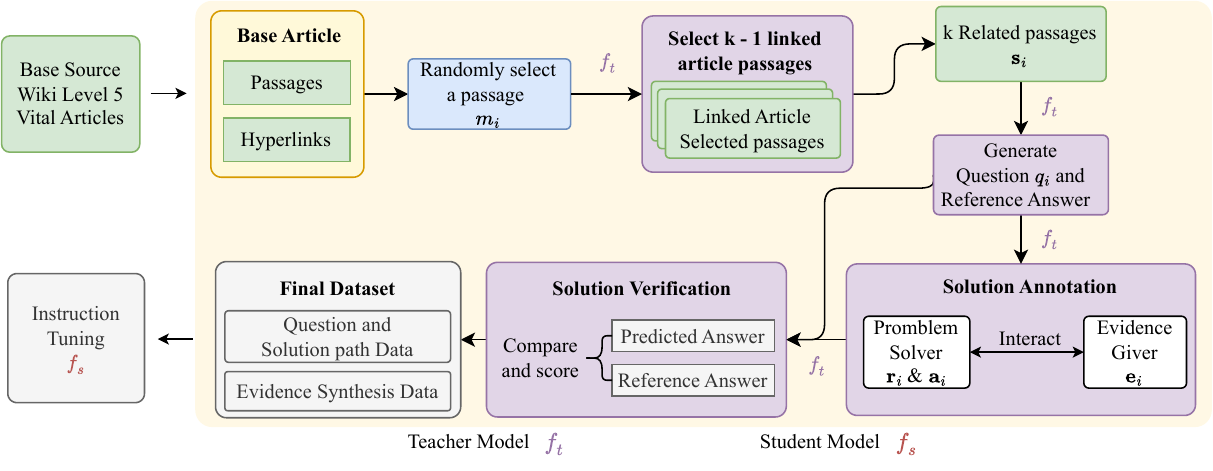} 
    \caption{Overview of the proposed SynAgent-RAG Dataset}
    \label{fig:data_pipeline}
\end{figure*}

\subsection{Working Memory Module}
The Working Memory module functions as a prompt memory, designed to store the input query and internal logs, including previous thoughts, actions generated by the LLM, and the extracted evidence obtained through the LLM's interactions with tools at steps. This memory is processed by the LLM to inform and guide subsequent actions. Furthermore, the Working Memory module ensures the system's transparency and explainability by recording the reasoning process, including the gathered knowledge and the decisions made at each step. This documentation provides insights into how conclusions were reached, enhancing trust and interpretability of the system.

\subsection{Agent Loop}
With all the defined modules, the agent operates within a loop that can be terminated either by the agent itself or when reaching a preconfigured computing budget. In the case of the agent, the entire pipeline terminates when the planning process confirms that sufficient external evidence has been gathered to answer the input query. In other cases, a parameter $k$ is preconfigured as the agent computing budget limit to limit the agent from processing too much information, and the loop is terminated when the computing budget is exhausted. Finally, in either case, the agent triggers a 'Final answer' action to aggregate all collected evidence and provide the final answer.

\section{SynAgent-RAG Dataset}
While the framework is fully compatible with bigger LLMs (e.g., GPT-4), deploying it with smaller LLMs necessitates an additional training process to maintain stability across each step. To address this challenge, we introduce SynAgent-RAG, a synthetic dataset designed for Agent-UniRAG. This is achieved through a distillation approach \cite{SemnaniYZL23}, where GPT-4 serves as the teacher model to generate data, and smaller models (e.g., LLama 3) are distilled versions. The primary objective of SynAgent-RAG is to empower the smaller LLM agent with the capability to reason, analyze, and synthesize information drawn from an external knowledge base before delivering a well-reasoned response to complex queries. The construction of SynAgent-RAG follows the process illustrated in Figure \ref{fig:data_pipeline}.

\subsection{Dataset Construction Process}
\subsubsection{Knowledge Base}
To construct an effective knowledge base for building the dataset that demands thoroughness, reliability, and up-to-date information across a wide range of fields, we utilized Wikipedia's Vital Articles Level 5\footnote{\url{https://en.wikipedia.org/wiki/Wikipedia:Vital_articles/Level/5}}. These articles represent a curated collection that encompasses essential topics for a comprehensive understanding of human knowledge. Prior to constructing the dataset, we carefully divided the articles into two separate sets: one for training and one for testing, to ensure a balanced evaluation of the model's performance.

\subsubsection{Question Generation}
To effectively generate questions that require multiple inference steps to arrive at a final answer, it is crucial to group related passages from source articles. We hypothesize that these related passages are interconnected through hyperlinks within each Wikipedia article. For each article, we randomly select a passage from the core content of the article as the main passage $m_i$. Then from passage $m_i$, to enhance the scalability of this process, we leverage GPT-4 to determine which hyperlinks are most relevant to the content of the main passage, following the prompt template (see Figure \ref{fig:extract_related_passages}). This process identifies up to 5 supporting articles with associated hyperlinks. Consequently, we obtain a set of main-supporting passage pairs $D_s = \{(m_i, \mathbf{s}_i)\}_{i=1}^n$.

Given the obtained set $D_s$, we construct both single and multi-hop questions $q_s$ that adhere to specific criteria following previous works in the field. Single-hop questions are designed to be straightforward, and answerable solely based on the information contained within the main passage $m_i$. In contrast, multi-hop questions necessitate information from multiple passages within the pair $\{(m_i, \mathbf{s}_i)\}$, requiring several inferential steps to derive the final answer. Furthermore, when employing GPT-4 with specified prompt templates (see Figure \ref{fig:gen_multihop_question} and \ref{fig:gen_singlehop_question}) the questions and long-form reference answers generated exhibit a high level of reasoning and analysis capability.

\subsubsection{Solution Annotation}
The solution annotation resulting from the planning and action decision of the teacher model to solve complex tasks is the key to effectively distilling the strong reasoning capabilities of student models. In this process, we generate solution annotations for questions that include a series of steps: \textit{Thought}, \textit{Action}, and \textit{Evidence Feedback}. Starting with the original question $q_i$, each step $t$, GPT-4 is required to perform two tasks replicating the real-world RAG scenario when the process of retrieving external knowledge is needed: i) provide a short rationale on how to utilize the Search Tool to address the question (Thought $r^t_i$) and formulate a detailed search query (Action $a^t_i$) to retrieve necessary information. ii, using the search query $a^t_i$ and the relevant sources $\{(m_i, \mathbf{s}_i)\}$ for the question $q_i$, GPT-4 will extract the most concise information from those sources and synthesize it as Evidence Feedback $e^t_i$. The results set at step $t$, comprising $\{r^t_i, a^t_i, e^t_i\}$, are concatenated with the question and prior steps in the order of $q_i, r^1_i, a^1_i, e^1_i, …,  r^t_i, a^t_i, e^t_i$ and used as the input for the agent to determine the plan and actions in the subsequent step $t+1$. The process continues until the agent concludes with a statement "I have the final answer" indicating sufficient evidence has been gathered. At this point, denoted as step $T$, the final answer is also provided. Finally, the solution annotation for the question $q_i$ includes thoughts $\mathbf{r}_i = \{r^1_i, …, r^T_i\}$, search queries $\mathbf{a}_i = \{a^1_i, …, a^{T-1}_i\}$, evidence feedbacks $\mathbf{e}_i = \{e^1_i, …, e^{T-1}_i\}$ and final answer. Details of prompts for the process are in Figure \ref{fig:extract_evidence} and \ref{fig:solution}.

\subsubsection{Annotation Verification}
Since the data are generated by an LLM, there are instances where the entire process may fail to provide the final answer. To address this, we implement both human and technical checks to ensure the scalability and reliability of the process. Additionally, we introduce an instruction eliminator, referred to as the Verification Module, to filter out failed annotations. We observe and hypothesize that if the process can produce a final answer similar to the reference answer then the annotation quality is considered high. Using a specified prompt template, GPT-4 is tasked with generating a brief rationale and then assigning an integer score ranging from 0 to 5, indicating the degree of similarity between the predicted answer and the reference answer and the relevancy to the input query. By employing the Solution Verification Module to filter annotations, we ensure the quality of the dataset by retaining only those annotations that achieve a score of 4 or 5.
\subsection{Dataset Analysis}
\begin{figure}[!h]
  \centering
  \includegraphics[width=0.9\columnwidth]{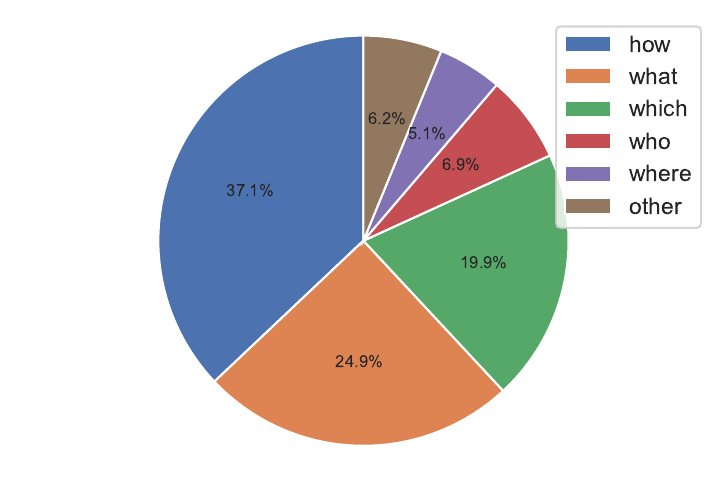}
  \caption{Question type distribution on the training set.}
  \label{fig:wh_question}
\end{figure}

After the annotation generation process, our dataset comprises 16,987 annotated training samples and 1,197 testing samples. Figure \ref{fig:wh_question} shows the distribution of question types and indicates that our dataset largely consists of 'how' questions confirming our initial goal of constructing a dataset to enhance the agent's ability to reason and synthesize information.
In real-world applications, RAG systems typically handle relatively simple queries. To account for this, we deliberately incorporated a higher proportion of queries requiring minimal search and fewer pieces of supporting evidence. On average, each training annotation in our dataset necessitates two supporting passages. This design ensures that our dataset reflects practical demands while accommodating varying complexities of user queries. Moreover, to the best of our knowledge, our dataset is the first dataset to integrate Chain of Thought (COT) reasoning, offering enhanced guidance for the agent to interact with external knowledge.

\section{Experiments}
\subsection{Experimental Setup}
\subsubsection{Datasets and Evaluation Metrics}
We evaluate the unified RAG model using both single-hop and multi-hop datasets. Specifically, we employ six benchmark datasets: three single-hop (SQuAD \cite{RajpurkarZLL16}, Natural Questions \cite{KwiatkowskiPRCP19}, TriviaQA \cite{JoshiCWZ17}) and three multi-hop (MusiQue \cite{TrivediBKS22}, HotpotQA \cite{Yang0ZBCSM18}, 2WikiMultiHopQA \cite{HoNSA20}). To compare our model against recent state-of-the-art RAG systems on these datasets, which feature short-form answers, we utilize F1, Exact Match (EM), and Accuracy (Acc) as evaluation metrics. The F1 score measures the overlap of words between the predicted and ground truth answers, EM checks for exact matches, and Acc verifies whether the predicted answer includes the ground truth. To adapt to the short-form answer setup, we use GPT-4 to extract concise answers from the detailed responses generated by the agent, as illustrated in Figure \ref{fig:extract_short_answer}. With each dataset, we benchmark on 500 samples per dataset processed by \cite{JeongBCHP24} and \cite{trivedi-etal-2023-interleaving}.

Additionally, we evaluate performance on the SynAgent-RAG test set, comparing small open-source LLMs with larger models utilized as the backbone for the Agent-UniRAG framework. We employ ROUGE-L and BLEU metrics to assess long-form answers. ROUGE-L, based on the longest common subsequence (LCS), measures similarity, while BLEU calculates n-gram precision, incorporating a brevity penalty to account for fluency and accuracy. Given the distinct response styles of each model, a comprehensive evaluation requires assessing their ability to exhibit analytical skills and produce logically coherent long-form responses. To this end, we also use GPT-Score, an LLM-based evaluator, which prompts an LLM to compare the generated answer with the reference and input queries.
 GPT-Score specifically evaluates the semantic alignment between the predicted and reference answers, thereby providing a more nuanced assessment of model performance.

\subsubsection{Retrieval System and Corpus Setup}
For the experiments on the short-form answer datasets, to ensure a fair comparison with the methodologies employed by \cite{JeongBCHP24}, we utilize the BM25 retriever as the baseline retrieval model across all corpus. In addition to the BM25 baseline retrieval model, we also experiment with adding the Multilingual E5 Large model \cite{wang2024multilingual} as the dense reranker after the sparse BM25 retrieval step to observe the effect of better retrieving results can lead to better agent performance. For the external corpus, we index different sources for different dataset types. Specifically, for single-hop datasets, follow \cite{JeongBCHP24} we use the Wikipedia corpus preprocessed by \cite{DBLP:journals/corr/abs-2004-04906}, while for multi-hop datasets, we use the corpus preprocessed by \cite{trivedi-etal-2023-interleaving}.

For the experiment on the test set of our SynAgent-RAG dataset, instead of indexing documents into a corpus, we focus on measuring the model's reasoning capability under optimal retrieval conditions. Here, we leave out the performance of retrieval systems and assume that the retrieved documents are correct and relevant to the original question by directly returning the reference documents as the results of the retrieval phase.

\subsubsection{Models}
\begin{table*}[!h]
  \centering
  \begin{adjustbox}{width=0.9\textwidth}
  \begin{tabular}{lccccccccccc}
    \hline
    \multirow{2}{*}{\textbf{Model}} &\multirow{2}{*}{\textbf{\makecell{Max\\Search}}} &  \multirow{2}{*}{\textbf{\makecell{Top K/\\BiEncoder}}} & \multicolumn{3}{c}{\textbf{SQUAD}} & \multicolumn{3}{c}{\textbf{Natural Question}}& \multicolumn{3}{c}{\textbf{TrivialQA}}\\
    &  &   &  EM & F1 & Acc& EM & F1 & Acc& EM & F1 & Acc\\
    \hline
    \hline
    Self-RAG*     &   No limit  &&1.6 & 11.9 & 20.8	&39.2 & 47.1 & 42.4	&14.6 & 33.7 &60.2\\
    IRCoT*     &   No limit  &&17.4 & 31.5 & 26.2	&35.6 & 49.7 & 57.8	&54.8 & 67.1 &68.0\\
    Adaptive-RAG*    &  No limit & &18.0 & 33.8 & 29.2&	32.4& 46.8 & 54.8	&55.2 &66.5 & 65.8       \\
    \hline
    Agent-UniRAG    &  1 & 8 / No& 23.8 & 34.5 & \textbf{49.6}&43.4 & 51.6 & 61.2	& 57.6 & 65.8& \textbf{71.2}\\
    Agent-UniRAG    &  1 & 12 / No&  26.6 & 38.1 & 48.6	& 45.2 & 53.9 & 61.2	& 57.0 &66.2 & 69.0 \\
    Agent-UniRAG    &  No limit & 8 / No & 26.4& 38.6&  42.2	&  45.8 &  55.3&  57.6	&  58.6 &  66.7&  70.0  \\
    Agent-UniRAG    &  No limit & 12 / No& 28.2 & 40.8 & 42.2	& 48.0 & 57.3&58.8& 57.4 & 67.2 & 67.4\\
    Agent-UniRAG    &  No limit & 12 / Yes& \textbf{32.8} &\textbf{46.9} & 42.8	& \textbf{59.2}& \textbf{68.6}& \textbf{64.6}& \textbf{63.6}& \textbf{72.5} & 71.0 \\
    \hline
  \end{tabular}
  \end{adjustbox}
  \caption{Results on different single-hop benchmark datasets. * are taken from \citet{JeongBCHP24} with GPT-3.5 as the backbone LLM for both previous approaches. Bold texts indicate the best results.}
  \label{tab:singlehop_results}
\end{table*}
\begin{table*}[!h]
\centering
\begin{adjustbox}{width=0.9\textwidth}
  \begin{tabular}{lccccccccccc}
    \hline
    \multirow{2}{*}{\textbf{Model}} &\multirow{2}{*}{\textbf{\makecell{Max\\Search}}} &  \multirow{2}{*}{\textbf{\makecell{Top K/\\BiEncoder}}} & \multicolumn{3}{c}{\textbf{MuSiQue}} & \multicolumn{3}{c}{\textbf{HotpotQA}}& \multicolumn{3}{c}{\textbf{2WikiMultiHopQA}}\\
    &  &   &  EM & F1 & Acc& EM & F1 & Acc& EM & F1 & Acc\\
    \hline
    \hline
    Self-RAG*     &   No limit  &&1.2 & 8.2 & 11.8	&5.6 & 17.8 & 30.6	&3.0 & 19.1 &39.0\\
    IRCoT*     &   No limit  &&23.0 & 32.5 &31.6&	45.8&58.3 &52.2&	52.2&66.0 & \textbf{62.4}\\
    Adaptive-RAG*   &  No limit && 21.8 & 32.6 & 29.6&	40.4 &52.5 & 47.0	&46.6 & 60.0 & 56.8     \\
   \hline
    Agent-UniRAG    &  No limit & 8 / No& 26.4 &35.2& 27.8	& 47.6& 56.2& 48.8& \textbf{60.2}& \textbf{66.7}& \textbf{61.8}  \\
    Agent-UniRAG    &  No limit & 12 / No& 26.2 & 35.3& 28.2& 48.6& 58.2& 50.6& 59.8 &66.6& 61.8	 \\
    Agent-UniRAG    &  No limit & 12 / Yes&  \textbf{30.4} & \textbf{39.8}& \textbf{32.2}& \textbf{50.2}& \textbf{59.9}& \textbf{52.4}	& 58.4 & 64.9& 60.6 \\
    \hline
    w/o Evidence Reflector & No Limit & 12 / No & 20.2 & 29.9 & 21.4 & 49.4 & 59.9 & 52.2 & 51.2 & 57.94 & 53.2\\
    w/o Planning & 1 &  12 / No  & 10.2 & 15.5 & 11.4 & 37.4 & 43.2 & 37.6 & 36.8 & 43.5 & 37.6\\
    \hline
  \end{tabular}
  \end{adjustbox}
  \caption{Results on different multi-hop benchmark datasets. * are taken from \citet{JeongBCHP24} with GPT-3.5 as the backbone LLM for both previous approaches. Bold texts indicate the best results.}
  \label{tab:multihop_results}
\end{table*} 

In this study, we compare our approach, Agent-UniRAG, against several retrieval-augmented LLM strategies, including \textit{Self-RAG} \cite{abs-2310-11511} which adaptively retrieves passages on-demand, and generates and reflects on retrieved passages, \textit{Adaptive-RAG} \cite{JeongBCHP24}, which dynamically adjusts retrieval based on question complexity, and \textit{IRCoT} \cite{trivedi-etal-2023-interleaving}, a state-of-the-art method leveraging iterative retriever and LLM interaction through Chain-of-Thought In-Context reasoning. The baseline models in these methods utilize GPT-3.5-Turbo, known for its larger size compared to our approach, which is based on LLama-3-Instruct. To further assess the effectiveness of our framework, we conducted an ablation study on multihop datasets. First, we removed the \textbf{Reflector Module} to assess the impact of directly utilizing the retrieved knowledge, which may include noise, as evidence feedback for the agent, whether it will lead to degradation in the performance. Second, we evaluated the effect of bypassing the gradual retrieval process by removing the \textbf{Planning Module}. In this scenario, the LLM was tasked with generating all necessary queries first, subsequently using the retrieved information to directly answer the input query. This setup helps understand the importance of iterative information retrieval in enhancing the agent's decision-making accuracy.

\subsubsection{Training Configurations}
Agent-UniRAG uses the instruction version of Meta-Llama-3-8B\footnote{https://huggingface.co/meta-llama/Meta-Llama-3-8B-Instruct} as the backbone open-source LLM model, and fine-tune instruction on the proposed SynAgent-RAG dataset. The fine-tuning process spanned 10 hours on a single DGX node with 8 A100 GPUs, each equipped with 40GB of VRAM. The learning rate was set at $2e^{-5}$, and the global batch size was 256. The model was trained for 2 epochs using the AdamW optimizer \cite{adamw}.

\subsubsection{Training Prompt Template}
We distill Agent-UniRAG in a multi-task setting by fine-tuning three subtasks following the proposed framework to guide its planning, action decisions, and filter evidence feedback. Annotations are organized in conversational formats to facilitate interaction between components, which include:
\\
\textbf{Conversation planning module annotation}:
As illustrated in Figure \ref{fig:train_solve}, we start by using the user role to provide the question $q$ in the prompt. The planning module then appends the first thought $r^1$ and the initial search query $a^1$ as the first response supervision. For subsequent turns, we act as the user and provide the extracted evidence $e^{t-1}$ of the last search query $a^{t-1}$ to the planning module. The response supervision dictates whether the planning should terminate by the thought \textit{“I have the final answer.”}; if not, the response should include a new thought $r^t$ along with a new search query $a^t$.
\\
\textbf{Conversation final answer annotation}:
Instead of letting the LLM generate the final answer in the planning module as in the data generation process, we want to add more control to the pipeline by separating the task of providing the final answer to a subtask. In that way, we collect the gathered evidence $\{e^1, …, e^{T-1}\}$ and provide the question as the user prompt and treat the final answer as the response (depicted in Figure \ref{fig:train_final_answer}).
\\
\textbf{Conversation Evidence Reflector annotation}:
As shown in Figure \ref{fig:train_extract_evidence}, we provide the search query $a^t$ and the relevant source containing the main-supporting passages pair $\{m, \mathbf{s}\}$ corresponding to the user turn. All the extracted evidence $e^t$ serves as the user's prompt response.

Since SynAgent-RAG annotations are conversational, we structure them as $\{x_1, y_1, \ldots, x_i, y_i, \ldots, x_n, y_n\}$, where $x_i$ is the $i$-th user prompt and $y_i$ indicates its responses. During training, we input each entire multi-turn annotation into the model, calculating loss solely on the tokens of the responses $Y = \{y_1, \ldots, y_i, \ldots, y_n\}$ and applying binary masking on the user prompt tokens to prevent computing loss on them. The final loss function is

\begin{equation}
  L = -\sum_j \log p_\pi(t_j \mid t_{<j}) \times {1}(t_j \in Y)
\end{equation}
where $t_j$ denotes the $j$-th input token and ${1}(\cdot)$ is a Boolean indicator function.
\label{sec:appendix3}

\subsection{Main Results}
We present a detailed performance comparison of the proposed approach with previous methods, as shown in Table \ref{tab:singlehop_results} for single-hop datasets and Table \ref{tab:multihop_results} for multi-hop datasets. Notably, Agent-UniRAG, which leverages a small open-source LLM as its backbone, demonstrates competitive performance relative to recent state-of-the-art models that utilize significantly larger LLMs. A key strength of our model is its ability to handle diverse query types uniformly and simultaneously. In addition to that, we have three specific observations.\\
\textbf{\uline{Agent-UniRAG can effectively interact with the external knowledge base.}}
We observe that increasing the search limits and the number of top-K retrieved documents leads to performance improvements. Specifically, with the top-K retrieval set to 12 and the integration of a dense encoder module for reranking, our proposed Agent-UniRAG substantially outperforms previous methods, achieving state-of-the-art results on the majority of benchmark RAG datasets in this research field. Additionally, in the single-hop settings, when the maximum search limit is increased from 1 to 'No limit' we observe a further increase in performance, highlighting the LLM agent's capability to interact with evidence feedback and then reason and refine search queries to gather better retrieval results.
\\
\textbf{\uline{The importance of designed modules in the pipeline.}}
In the multi-hop reasoning setting, with the conducted ablation studies to assess (Table \ref{tab:multihop_results}), removing the Evidence Reflector module resulted in noticeable performance degradation, particularly in more complex datasets like MuSIQue \cite{TrivediBKS22}, underscoring the critical role of the Evidence Reflector in provisioning concise and relevant evidence feedback to help the agent make better subsequent decisions. We also removed the Planning module, which serves as the central component of the pipeline. The removal of this module led to a more substantial decline in performance metrics, thereby illustrating its pivotal role in orchestrating the agent’s multi-step reasoning process and the necessity of iterative information retrieval.
\begin{table}[!h]
\begin{adjustbox}{width=\columnwidth}
 \begin{tabular}{llccc}
    \hline
    \textbf{Agent LLM}  &  \textbf{Rouge-L} &  \textbf{BLEU} & \textbf{GPT-Score} & \textbf{Step}\\
    \hline
    \hline
    Llama-3-70B-Inst   &  0.36  &  0.12 & 3.62 & 5.68\\
    GPT-4-Turbo          &   0.35   & 0.13   & 4.35 & 2.27\\
    \hline
    \textbf{Agent-UniRAG}     &  0.36    & 0.15   &  4.19 & 2.08\\
    \hline
  \end{tabular}
  \end{adjustbox}
  \caption{Agent-UniRAG in compare with LLama-3-70B-Inst and GPT-4-Turbo on SynAgent-RAG test set.}
  \label{tab:agent}
\end{table}
\\
\textbf{\uline{Effectiveness of SynAgent-RAG dataset in distilling the reasoning capability.}}
Table \ref{tab:agent} presents the results on the test set of SynAgent-RAG datasets. Upon analysis, it becomes evident that traditional metrics like Rouge-L and BLEU, focused on the lexical overlap, is insufficient for evaluating the reasoning and accuracy in long-form answers. In contrast, GPT-Score, leveraging LLMs for semantic evaluation, provides a more accurate assessment. As a result, our proposed Agent-UniRAG model, which is finetuned on the SynAgent-RAG training set, demonstrates strong performance based on GPT-Score, achieving comparable results to significantly larger models such as LLaMA-70B-Inst and GPT-4. Notably, Agent-UniRAG achieves this level of performance while utilizing fewer external search queries, highlighting its computational efficiency compared to more demanding computational resources typically required by larger LLMs and as an efficient solution for generating accurate long-form answers. This result also underscores the effectiveness of the SynAgent-RAG dataset in distilling reasoning capabilities from a larger LLM (GPT-4) into a more compact framework.

\section{Conclusion}
Most previous works for LLM RAG tasks either handle single-hop queries with unnecessary computational complexity or fail to address complex multi-hop queries. This study, inspired by the emerging LLM agent technologies, presents a novel approach for unified RAG systems with enhanced effectiveness and interpretability. Furthermore, we introduce SynAgent-RAG, a synthetic dataset that enables trainable open-source LLM agents for the unified RAG task. Compared with previous works using larger LLMs, the experiment shows promising results of the Agent-UniRAG with a small backbone (i.e., Llama-3-8B).
\section*{Limitations}
Agent-UniRAG, the unified Retrieval-Augmented Generation (RAG) model, has shown promising results in managing different types of queries including single-hop and multi-hop queries. Notably, this approach can be applied to open-source small language models (LLMs). However, real-world input often encompasses scenarios when queries are often more than just single-hop or multi-hop queries requiring access to external knowledge bases, they may also include non-RAG tasks such as creative writing or function calling. Therefore, a crucial direction for future research is to extend the proposed approach to handle broader types of queries, including both RAG and non-RAG types. Additionally, similar to other LLM agent architectures, Agent-UniRAG requires multiple calls to the language model to generate a final response, which introduces a computational challenge during inference. Consequently, optimizing LLM inference for agent architectures is another critical aspect of our future work.

\bibliography{custom}

\appendix

\section{Appendix}
\label{sec:appendix}
\section{Illustrations of Agent-UniRAG Sample}
\label{sec:data_illustration}
Tables \ref{tab:example1} and \ref{tab:example2} illustrate our data with examples, including the question, thought process, actions taken, evidence gathered, and the final long-form answer.
\begin{table*}[!h]
  \centering
  \begin{adjustbox}{width=\textwidth}
  \begin{tabular}{lc}
    \hline
    \textbf{Question} & \textbf{Agent-UniRAG Output} \\
    \hline
    \hline
    \makecell[l]{What highway was\\ renamed  in honor of \\ Tim Russert?}  & \makecell[l]{
    \#\#\# \textbf{Thought}: The first step is to use the search\_engine tool to find which highway \\ was renamed in honor of Tim Russert.
    \\
    \#\#\#  \textbf{Action - Search Input}: highway renamed in honor of Tim Russert
    \\
    \#\#\# \textbf{Evidence}: On July 23, 2008, U.S. Route 20A leading to the  Buffalo Bills' \\ Ralph Wilson Stadium in Orchard Park, New York was renamed the 'Timothy J.\\ Russert Highway' in honor of Tim Russert.
    \\
    \#\#\# \textbf{Thought}: I have the final answer
    \\
    \#\#\# \textbf{Action - Final Answer}: U.S. Route 20A was renamed the 'Timothy  J. Russert \\  Highway' in honor of Tim Russert.
    }
    \\
    \hline
  \end{tabular}
  \end{adjustbox}
  \caption{Example of Agent-UniRAG output with single-hop query}
  \label{tab:example1}
\end{table*}
\begin{table*}[!h]
  \centering
  \begin{adjustbox}{width=\textwidth}
  \begin{tabular}{lc}
    \hline
    \textbf{Question} & \textbf{Agent-UniRAG Output} \\
    \hline
    \hline
    \makecell[l]{How do Jerry Falwell's \\beliefs about the \\Antichrist as a specific\\ person contrast with  \\Martin Wight's  \\interpretation of  \\the Antichrist concept\\ after World War II?}  & \makecell[l]{
    \#\#\# \textbf{Thought}: Understand Jerry Falwell's beliefs about the Antichrist as a \\ specific person.
    \\
    \#\#\#  \textbf{Action - Search Input}: Jerry Falwell beliefs about the Antichrist
    \\
    \#\#\# \textbf{Evidence}: In 1999, Jerry Falwell professed that the Antichrist would likely\\ emerge within a decade, with the prediction that this figure would be Jewish,\\ which led to allegations of anti-Semitism and an ensuing apology from Falwell.\\ He believed the Antichrist and  Christ would share many attributes.
    \\
    \#\#\# \textbf{Thought}: Explore Martin Wight's interpretation of the Antichrist concept\\ after World War II to contrast with Falwell's views.
    \\
    \#\#\# \textbf{Action - Search Input}: Martin Wight interpretation of the Antichrist \\concept after World War II.
    \\
    \#\#\# Evidence: Post-World War II, Christian and political theorist Martin Wight \\interpreted the Antichrist not as an individual, but as a symbol of 'demonic \\concentrations of power' recurring over time.
    \\
    \#\#\# \textbf{Thought}: I have the final answer
    \\
    \#\#\# \textbf{Action - Final Answer}: Jerry Falwell's beliefs about the Antichrist focused \\on the idea that the Antichrist would be a specific individual, with Falwell \\ predicting in 1999 that this person would likely emerge within a decade and \\possess similarities to Christ. In contrast, Martin Wight's post-World War II \\ interpretation of the Antichrist was symbolic, referring to it as 'demonic \\concentrations of power' that recur throughout history, rather than an individual \\figure.
    }
    \\
    \hline
  \end{tabular}
  \end{adjustbox}
  \caption{Example of Agent-UniRAG output with multi-hop query}
  \label{tab:example2}
\end{table*}

\section {Prompt Templates}
\label{prompt_template}
\begin{figure*}[!h]
    \centering 
    \includegraphics[width=\textwidth]{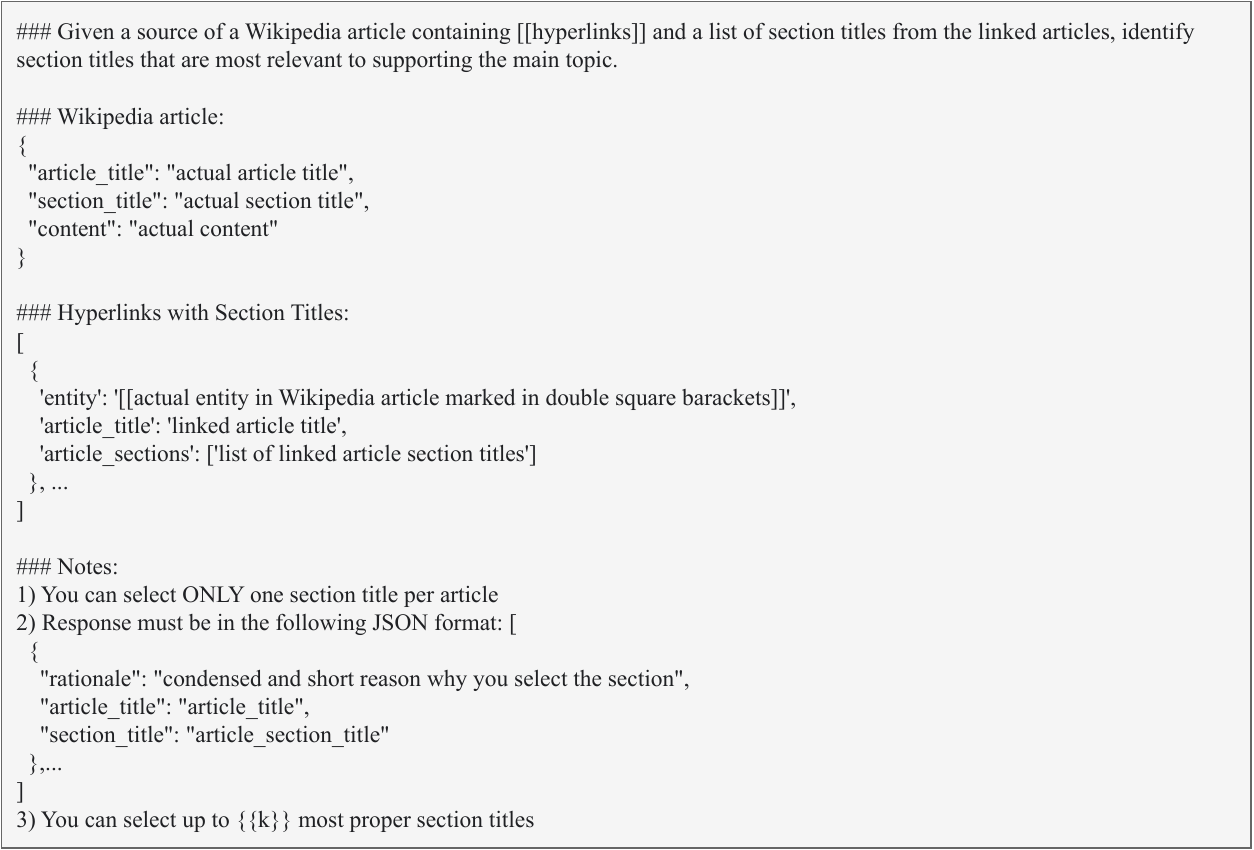} 
    \caption{Prompt template for GPT4 to find related section content from articles.}
    \label{fig:extract_related_passages}
\end{figure*}

\begin{figure*}[!h]
    \centering 
    \includegraphics[width=\textwidth]{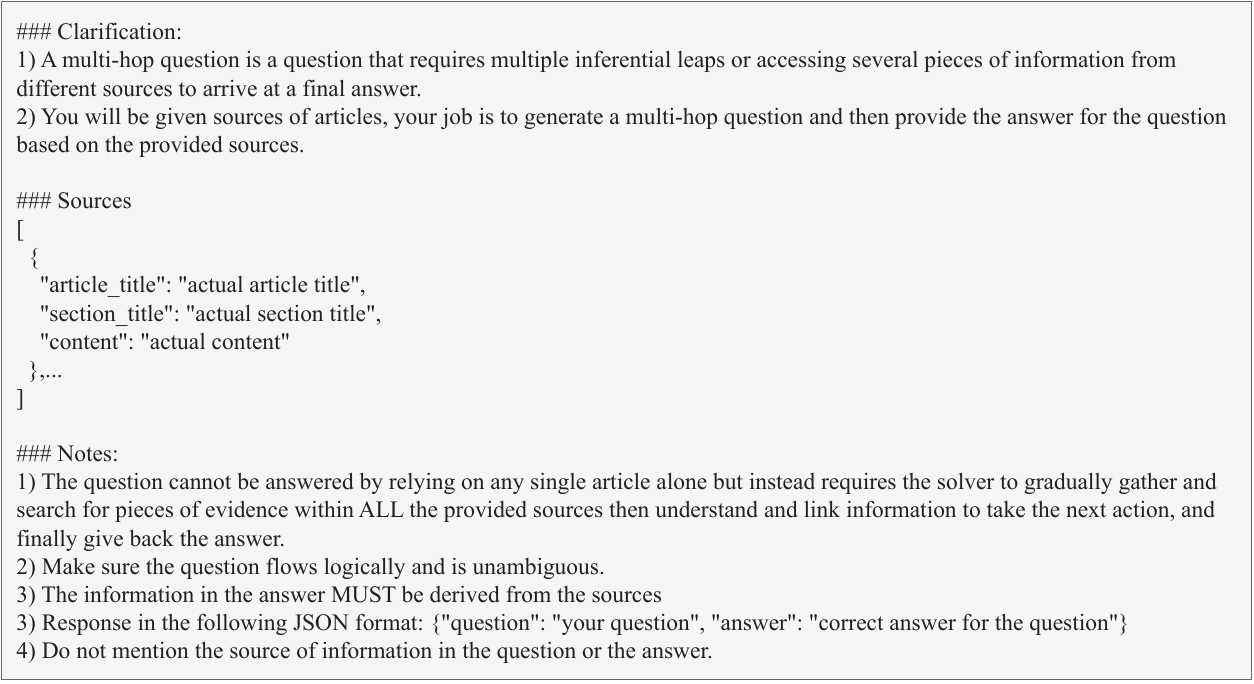} 
    \caption{Prompt template for GPT4 to generate multi-hop questions.}
    \label{fig:gen_multihop_question}
\end{figure*}

\begin{figure*}[!h]
    \centering 
    \includegraphics[width=\textwidth]{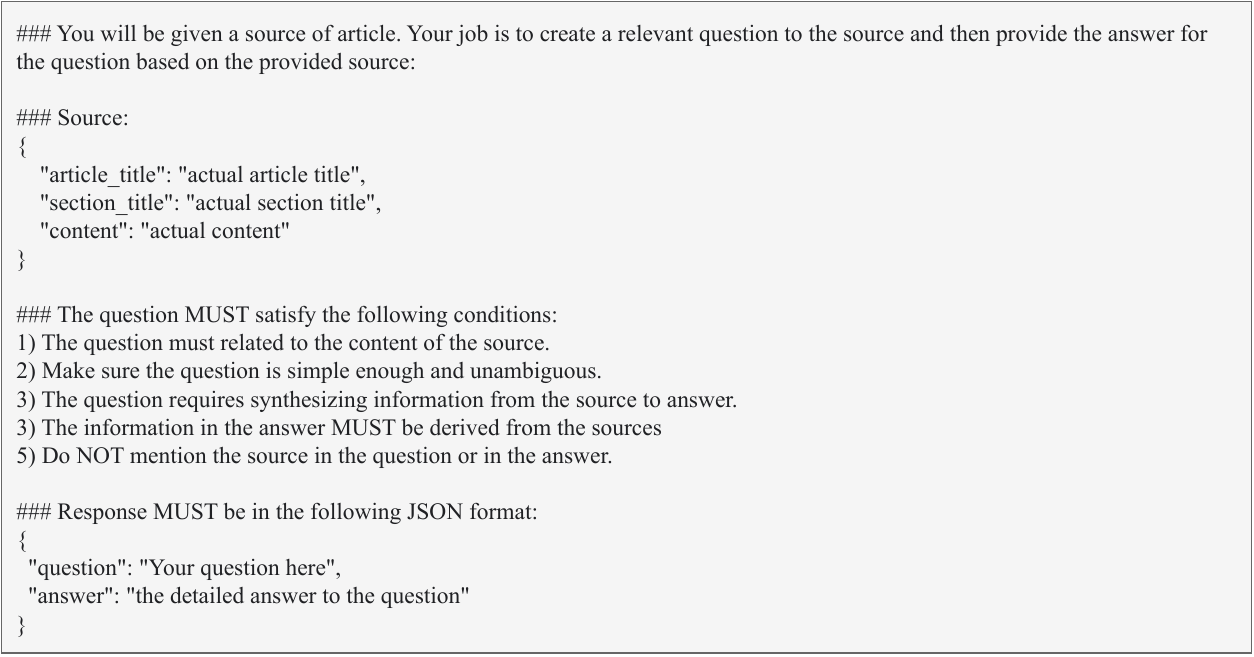} 
    \caption{Prompt template for GPT4 to generate single-hop questions.}
    \label{fig:gen_singlehop_question}
\end{figure*}

\begin{figure*}[!h]
    \centering 
    \includegraphics[width=\textwidth]{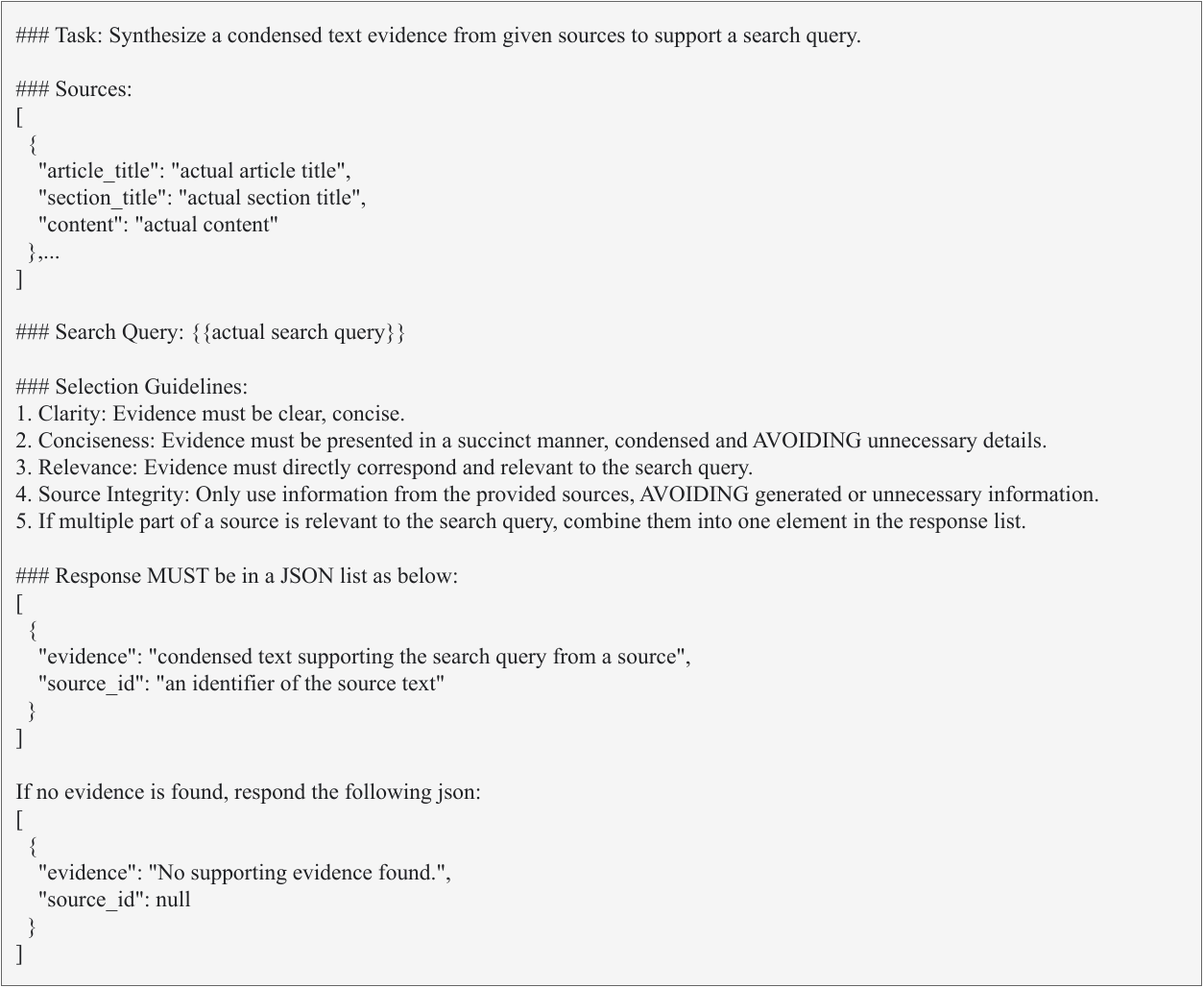} 
    \caption{Prompt template for GPT4 to extract evidence from a list of sources}
    \label{fig:extract_evidence}
\end{figure*}

\begin{figure*}[!h]
    \centering 
    \includegraphics[width=\textwidth]{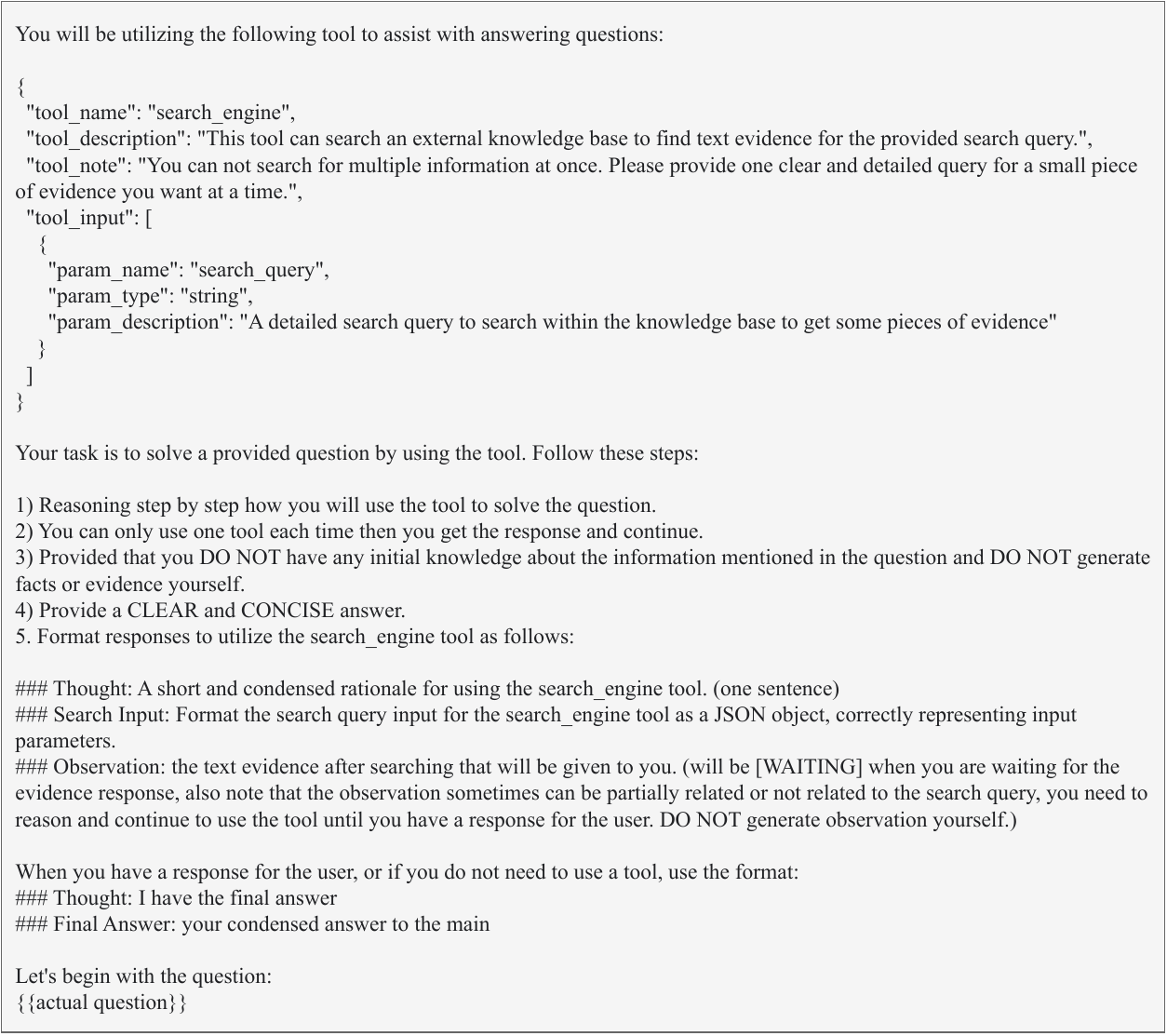} 
    \caption{Prompt template for GPT4 to reason, use tools and provide the final answer for a question}
    \label{fig:solution}
\end{figure*}

\begin{figure*}[!h]
    \centering 
    \includegraphics[width=\textwidth]{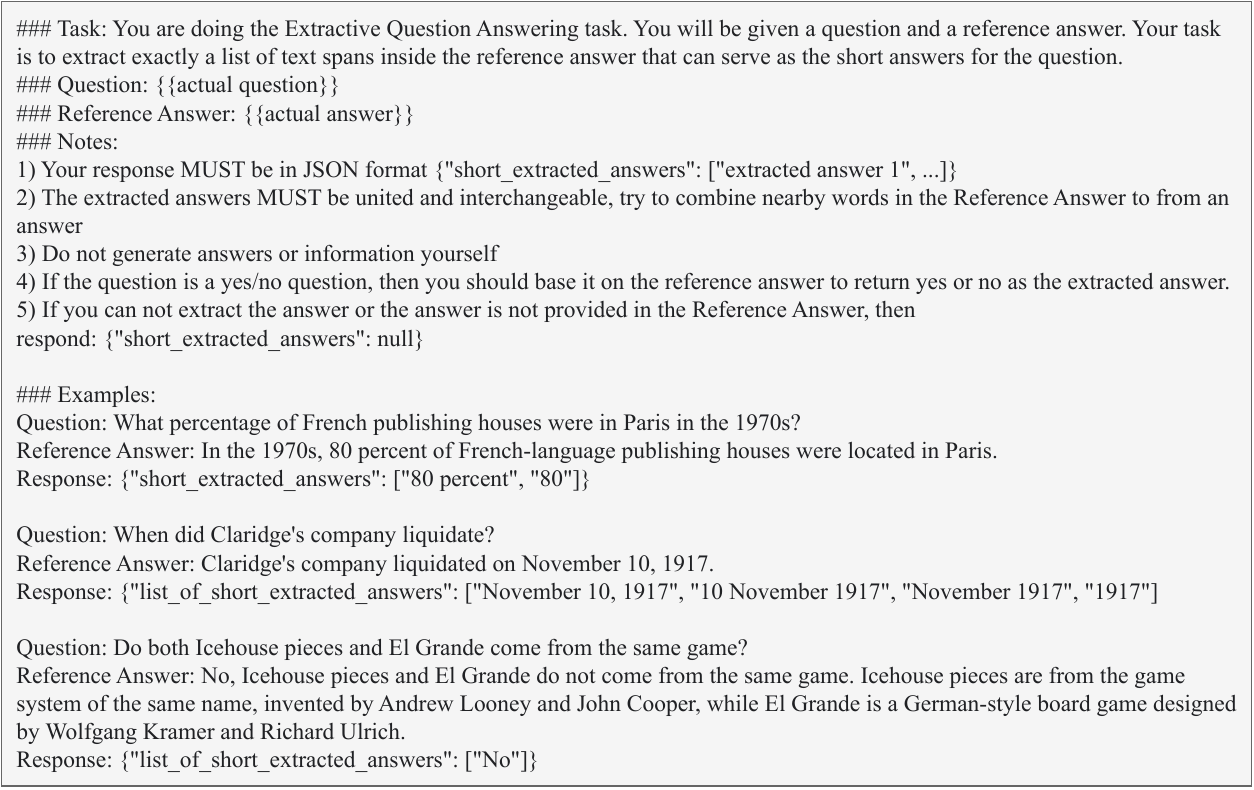} 
    \caption{Prompt template to extract short answer from long answer}
    \label{fig:extract_short_answer}
\end{figure*}

\begin{figure*}[!h]
    \centering 
    \includegraphics[width=\textwidth]{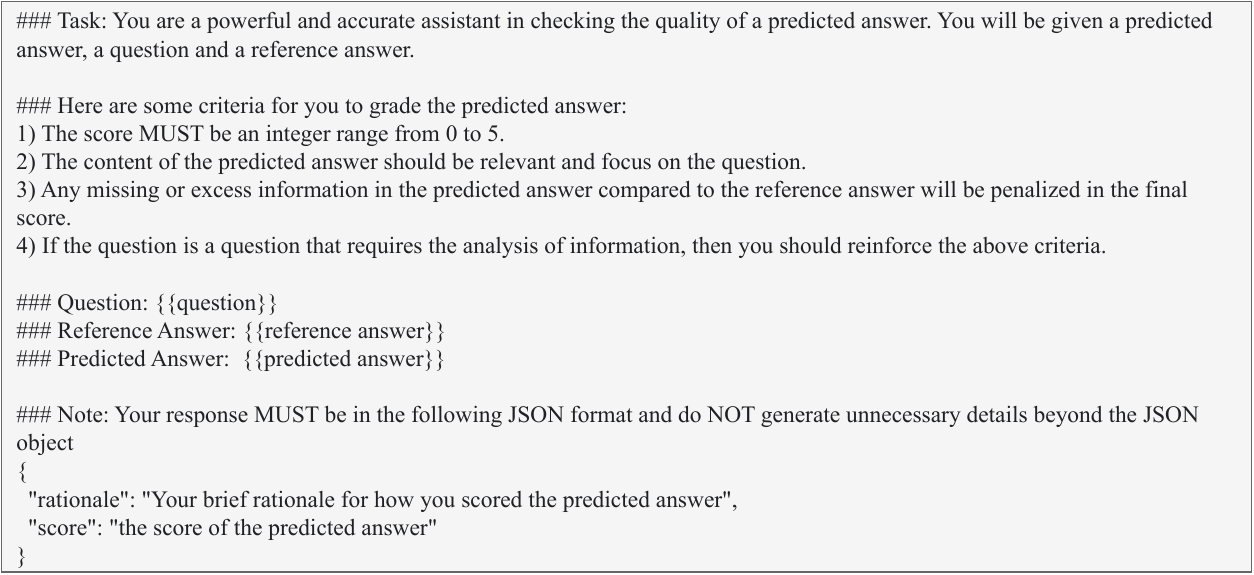} 
    \caption{Prompt template for GPT4 to compare and score the predicted answer and the reference answer.}
    \label{fig:gpt_score}
\end{figure*}

\begin{figure*}[!h]
    \centering 
    \includegraphics[width=\textwidth]{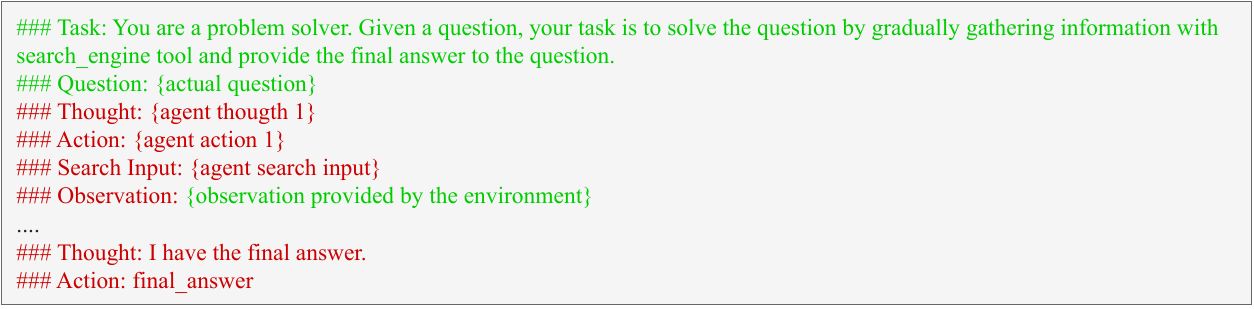} 
    \caption{Training prompt template for Agent-UniRAG to reason and use tools. Loss is computed only on the red part as the GPT turns in the conversation setup.}
    \label{fig:train_solve}
\end{figure*}

\begin{figure*}[!h]
    \centering 
    \includegraphics[width=\textwidth]{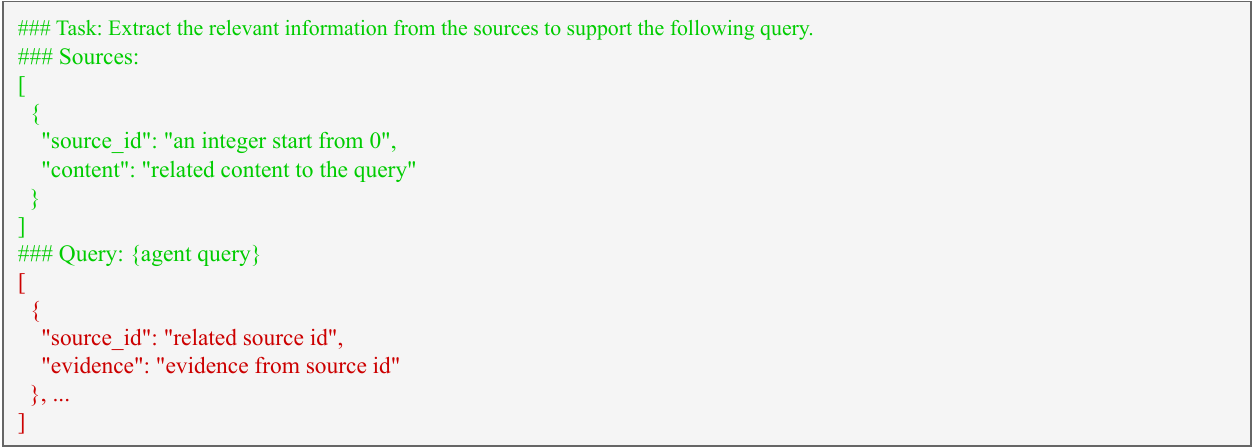} 
    \caption{Training prompt template for Agent-UniRAG extract related evidence for a query from sources of content. Loss is computed only on the red part as the GPT turns in the conversation setup.}
    \label{fig:train_extract_evidence}
\end{figure*}

\begin{figure*}[!h]
    \centering 
    \includegraphics[width=\textwidth]{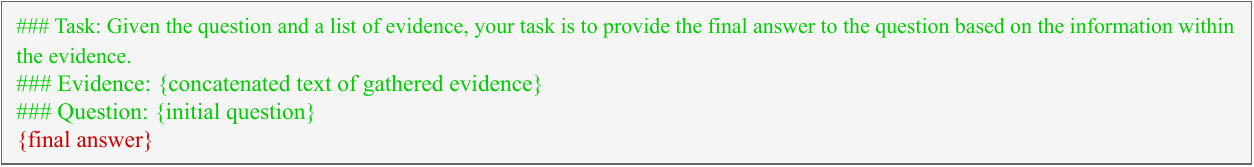} 
    \caption{Training prompt template for Agent-UniRAG to provide final answer from gathered evidence. Loss is computed only on the red part as the GPT turns in the conversation setup.}
    \label{fig:train_final_answer}
\end{figure*}

For better reproducibility, we present all prompt templates in the appendix. Below is a quick reference list outlining the prompt templates and their usages:
\begin{itemize}
    \item Figure \ref{fig:extract_related_passages}: Find related sections from linked articles to the main article.
    \item Figure \ref{fig:gen_multihop_question}: Generate a multi-hop question from a cluster of related sections.
    \item Figure \ref{fig:gen_singlehop_question}: Generate a singlehop question from a section.
    \item Figure \ref{fig:extract_evidence}: Extract related evidence for a search query from a list of source content.

    \item Figure \ref{fig:solution}: Generate solution annotation for a question.
      \item Figure \ref{fig:extract_short_answer}: Extract short form answer from a long form answer.
    \item Figure \ref{fig:gpt_score}: Take GPT Score of an annotated answer to the reference answer.
    \item Figure \ref{fig:train_solve}: Training prompt template for the agent to reason and use tools.
    \item Figure \ref{fig:train_extract_evidence}: Training prompt template for the agent to extract related evidence for a query from sources of content.
    \item Figure \ref{fig:train_final_answer}: Training prompt template for the agent to provide the final answer for the question from gathered evidence
  
\end{itemize}
All prompts are zero-shot, except for the prompt in Figure \ref{fig:extract_short_answer}, which uses few-shot demonstrations to better guide the LLM to perform the task. These prompts were chosen because they perform effectively in practice.
\label{sec:figures}


\end{document}